\documentclass[sigconf,authorversion]{acmart}  
\usepackage{mystyle}
\newacronym{ace}{ACE}{adaptive calibration error}
\newacronym{aucpr}{AUC-PR}{area under the precision-recall curve}
\newacronym{aucroc}{AUC-ROC}{area under the receiver operating characteristic curve}
\newacronym{nll}{NLL}{negative log-likelihood}
\newacronym{ccs}{CCS}{Clinical Classifications Software}
\newacronym{ece}{ECE}{expected calibration error}
\newacronym{eicu}{eICU}{eICU Collaborative Research Database}
\newacronym{elbo}{ELBO}{evidence lower bound}
\newacronym{icu}{ICU}{intensive care unit}
\newacronym{kl}{KL}{Kullback-Leibler}
\newacronym{lstm}{LSTM}{long short-term memory}
\newacronym{mimic}{MIMIC-III}{Medical Information Mart for Intensive Care}
\newacronym{nn}{NN}{neural network}
\newacronym{ppv}{PPV}{positive predictive value}
\newacronym{pr}{PR}{precision-recall}
\newacronym{rnn}{RNN}{recurrent neural network}
\newacronym{vi}{VI}{variational inference}
\newacronym{ehr}{EHR}{electronic health record} 
\copyrightyear{2020}
\acmYear{2020}
\setcopyright{rightsretained}
\acmConference[ACM CHIL '20]{ACM Conference on Health, Inference, and Learning}{April 2--4, 2020}{Toronto, ON, Canada}
\acmBooktitle{ACM Conference on Health, Inference, and Learning (ACM CHIL '20), April 2--4, 2020, Toronto, ON, Canada}
\acmDOI{10.1145/3368555.3384457}
\acmISBN{978-1-4503-7046-2/20/04}

\begin{document}
\title{Analyzing The Role Of Model Uncertainty\\For Electronic Health Records}
\renewcommand{\shorttitle}{Analyzing The Role Of Model Uncertainty For Electronic Health Records}

\author{Michael W. Dusenberry}
\authornote{Work completed as a Google AI Resident.}
\authornote{Correspondence to: dusenberrymw@google.com.}
\affiliation{
  \institution{Google Brain}
}

\author{Dustin Tran}
\affiliation{
  \institution{Google Brain}
}

\author{Edward Choi}
\affiliation{
  \institution{Google Health}
}

\author{Jonas Kemp}
\affiliation{
  \institution{Google Health}
}

\author{Jeremy Nixon}
\affiliation{
  \institution{Google Brain}
}

\author{Ghassen Jerfel}
\affiliation{
  \institution{Google Brain, Duke University}
}

\author{Katherine Heller}
\affiliation{
  \institution{Google Brain}
}

\author{Andrew M. Dai}
\affiliation{
  \institution{Google Health}
}

\renewcommand{\shortauthors}{Dusenberry, et al.} \begin{abstract}
In medicine, both ethical and monetary costs of incorrect predictions can be significant, and the complexity of the problems often necessitates increasingly complex models.  Recent work has shown that changing just the random seed is enough for otherwise well-tuned deep neural networks to vary in their individual predicted probabilities.  In light of this, we investigate the role of model uncertainty methods in the medical domain. Using \gls*{rnn} ensembles and various Bayesian \glspl*{rnn}, we show that population-level metrics, such as \acrshort*{aucpr}, \acrshort*{aucroc}, log-likelihood, and calibration error, do not capture model uncertainty.  Meanwhile, the presence of significant variability in patient-specific predictions and optimal decisions motivates the need for capturing model uncertainty. Understanding the uncertainty for individual patients is an area with clear clinical impact, such as determining when a model decision is likely to be brittle. We further show that \glspl*{rnn} with only Bayesian embeddings can be a more efficient way to capture model uncertainty compared to ensembles, and we analyze how model uncertainty is impacted across individual input features and patient subgroups. \end{abstract}



\maketitle
\glsresetall

\section{Introduction}

Machine learning has found great and increasing levels of success in the last several years on many well-known benchmark datasets.  This has led to a mounting interest in non-traditional problems and domains, each of which bring their own requirements.  In medicine specifically, individualized predictions are of great importance to the field \citep{national2011toward}, and there can be severe costs for incorrect
decisions due to the risk to human life and associated ethical concerns  \citep{gillon1994medical}.

Existing state-of-the-art approaches using deep neural networks in medicine often make use of either a single model or an average over a small ensemble of models, focusing on improving the accuracy of probabilistic predictions \citep{harutyunyan2017multitask, rajkomar2018, xu2018raim, choi2018mime}. These works, while focusing on capturing the data uncertainty, do not address the \textit{model} uncertainty that is inherent in fitting deep neural networks \citep{kendall2017, malinin2018predictive}.  For example, when predicting patient mortality in an ICU setting, existing approaches might be able to achieve high \acrshort*{aucroc}, but will be unable to differentiate between patients
for whom the model is \textit{certain} about its probabilistic prediction, and those for whom the model is fairly \textit{uncertain}.

In this paper, we examine the use of model uncertainty specifically in the context of predictive medicine. Model uncertainty has made many methodological advances in recent years---including reparameterization-based variational Bayesian neural networks \citep{blundell2015,fortunato2017,kucukelbir2017automatic,louizos2017multiplicative}, Monte Carlo dropout \citep{gal2016dropout}, deep ensembles and efficient alternatives \citep{lakshminarayanan2017, wen2020}, and function priors \citep{hafner2018reliable,garnelo2018neural,malinin2018predictive}. 
Deep neural networks combined with advanced model uncertainty methods can directly impact clinical care by answering several questions that naturally occur in predictive medicine:

\begin{itemize}
    \item How do the realized functions in any of the approaches, such as individual models in the ensemble approach, compare in terms of population-level metric performance such as AUC-PR, AUC-ROC, or log-likelihood?
    \vskip 0.3em
    \item If and how does model uncertainty assist in calibrating predictions?
    \vskip 0.3em
    \item How does model uncertainty change across different patient subgroups, in terms of ethnicity, gender, age, or length of stay?
    \vskip 0.3em
    \item How do various feature values contribute towards model uncertainty?
    \vskip 0.3em
    \item How does model uncertainty affect optimal decisions made under a given clinically-relevant cost function?
\end{itemize}

\textbf{Contributions.\,}
Using sequence models on the \acrshort*{mimic} \citep{johnson2016} and eICU \cite{pollard2018eicu} clinical datasets, we make several important findings.  For the ensembling approach of quantifying model uncertainty, we find that the models within the ensemble 
can collectively exhibit a wide variability in predicted probabilities for some patients, despite being well-calibrated and having \textit{nearly identical dataset-level metric performance}.
We find that this even extends into the space of optimal decisions.  That is, models with nearly equivalent metric performance can disagree significantly on the final decision, thus transforming an "optimal" decision into a random variable.
Significant variability in patient-specific predictions and decisions can be an indicator of when a model decision is likely to be brittle, and we show that using a single model or an average over models can mask this information. This motivates the importance of model uncertainty for clinical decision systems.
Given this, we proceed with an analysis over different clinical tasks and datasets, looking at how model uncertainty is impacted across individual input features and patient subgroups. We then show that models with Bayesian embeddings can be a more efficient way to capture model uncertainty compared to deep ensembles. \section{Background}

\subsection{Data Uncertainty}
Data uncertainty can be viewed as uncertainty regarding a given outcome due to incomplete information, and is also known as ``output uncertainty'', ``noise", or ``risk'' \citep{knight1957}. This uncertainty is represented by the predictive distribution
\begin{equation}
    y \sim p(y | \mathbf{x})
\end{equation}
for the outcome $y$ given inputs $\mathbf{x}$. In a learning scenario, we could define a function $f(\mathbf{x}, \mathbf{w})$ with learnable parameters $\mathbf{w}$ that outputs a parameterization of the predictive distribution $p(y | \mathbf{x}, \mathbf{w})$, which is now conditioned on $\mathbf{w}$.
For binary tasks, the predictive distribution equates to a Bernoulli distribution, which is parameterized by a single probability value.  More specifically, for the binary case, this can be described as
\begin{equation}
\begin{split}
\boldsymbol{\lambda} &= f(\mathbf{x}, \mathbf{w}) \\
y &\sim \operatorname{Bernoulli}(\boldsymbol{\lambda}),
\end{split}
\label{eq:data-uncertainty-binary}
\end{equation}
where the model $f$, as a function of the inputs $\mathbf{x}$ and parameters $\mathbf{w}$, outputs the parameter $\boldsymbol{\lambda}$ (a vector of length one) for the Bernoulli distribution representing the conditional distribution $p(y | \mathbf{x}, \mathbf{w})$ for the outcome $y$. For multiclass tasks, the predictive distribution takes the form of a Multinomial distribution with a single trial (and parameterized by a vector $\boldsymbol{\lambda}$), and for regression tasks, one could use a continuous distribution such as a Gaussian.

\subsection{Model Uncertainty}
\label{sec:model-uncertainty}
Model uncertainty can be viewed as uncertainty regarding the true function underlying the observed process \citep{bishop2006}.  For a learned function $f(\mathbf{x}, \mathbf{w})$ of inputs $\mathbf{x}$ and parameters $\mathbf{w}$, this uncertainty is represented by a distribution over functions \citep{bishop2006, wilson2019bayesian}
\begin{equation}
    f \sim \mathcal{G}(f)
\end{equation}
which is often induced by a distribution over the function parameters \citep{bishop2006, blundell2015, fortunato2017, wilson2019bayesian}
\begin{equation}
    \mathbf{w} \sim p(\mathbf{w}).
\end{equation}
Because different functions can yield different predictive distributions,
a distribution over functions leads to a distribution over predictive distributions, representing \textit{disagreement} due to model uncertainty.
We can see this more formally by defining a function $g_\mathbf{x}(\mathbf{w}) = f(\mathbf{x}, \mathbf{w})$ for a given input $\mathbf{x}$, and then viewing this as a change of variables from $\mathbf{w}$ to $\boldsymbol{\lambda}$,
\begin{equation}
\begin{split}
\boldsymbol{\lambda} &= g_\mathbf{x}(\mathbf{w}) = f(\mathbf{x}, \mathbf{w}) \\
\mathcal{P}(\boldsymbol{\lambda} \in \mathcal{A} \mid \mathbf{x}) &= \int_{\{\mathbf{w} | g_\mathbf{x}(\mathbf{w}) = \boldsymbol{\lambda} \in \mathcal{A}\}} p(\mathbf{w}) \d{\mathbf{w}} \\
p(\boldsymbol{\lambda} | \mathbf{x}) 
&= \frac{\d{}}{\d{\boldsymbol{\lambda}}} \int_{\{\mathbf{w} | g_\mathbf{x}(\mathbf{w}) \leq \boldsymbol{\lambda}\}} p(\mathbf{w}) \d{\mathbf{w}},
\end{split}
\label{eq:change-of-vars}
\end{equation}
where the distribution over $\mathbf{w}$ is transformed into a distribution over $\boldsymbol{\lambda}$ (conditioned on $\mathbf{x})$. 
Thus, there is an induced distribution over the parameters of the predictive distribution due to uncertainty in the function space. For binary tasks, this would equate to a distribution of plausible probability values for a Bernoulli distribution.

We can then write down the final, marginalized predictive distribution
\begin{equation}
\begin{split}
p(y | \mathbf{x}) &= \int p(y | \mathbf{x}, \mathbf{w}) p(\mathbf{w}) \d{\mathbf{w}} \\
&= \int p(y | \mathbf{x}, \boldsymbol{\lambda}) p(\boldsymbol{\lambda} | \mathbf{x}) \d{\boldsymbol{\lambda}} \\
\end{split}
\label{eq:predictive-dist}
\end{equation}
in two equivalent forms.  Importantly, by considering the distribution $p(\boldsymbol{\lambda} | \mathbf{x})$ before marginalizing, we can compute two quantities of interest: the expected value $\mathbb{E}_{\boldsymbol{\lambda} \sim p(\boldsymbol{\lambda} | \mathbf{x})}[\boldsymbol{\lambda} | \mathbf{x}]$ (which is used in the marginalization of equation \ref{eq:predictive-dist}), and a measure of \textit{disagreement} (or uncertainty due to model uncertainty) such as the variance
$\operatorname{Var}[\boldsymbol{\lambda} | \mathbf{x}]$. It is also important to note that the variance in the final, marginalized \textit{predictive distribution} $p(y | \mathbf{x})$ will include \textit{both} data uncertainty \textit{and} model uncertainty sources, but it is not possible to distinguish the two from that marginalized distribution alone (and thus $p(\boldsymbol{\lambda} | \mathbf{x})$ is needed).

For the remainder of the paper, we will use the phrase \textit{predictive uncertainty distribution} to refer to the distribution $p(\boldsymbol{\lambda} | \mathbf{x})$ over the parameter(s) of the predictive distribution as induced by the uncertainty over model parameters.

\subsection{Calibration}
A model is said to be perfectly calibrated if, for all examples for which the model produces the same prediction $p$ for some outcome, the percentage of those examples truly associated with the outcome is equal to $p$, across all values of $p$.
If a model is systematically over- or under-confident, it can be difficult to reliably use its predicted probabilities for decision making.
The \gls*{ece} metric \citep{naeini2015} is one tractable way to approximate the calibration of a model given a finite dataset.  ECE computes a weighted average of the calibration error across bins, and is defined as
\begin{equation*}
\operatorname{ECE} = \sum_{b=1}^{B} \frac{n_b}{N} \left| \operatorname{acc}(b) - \operatorname{conf}(b) \right|,
\end{equation*}
where $n_b$ is the number of predictions in bin $b$, $N$ is the total number of data points, and $\operatorname{acc}(b)$ and $\operatorname{conf}(b)$ are the accuracy and confidence of bin $b$, respectively.
Recent work \cite{guo2017} has shown that modern deep \glspl*{nn} tend to be poorly calibrated.

\begin{figure}
\centering
\includegraphics[width=\columnwidth]{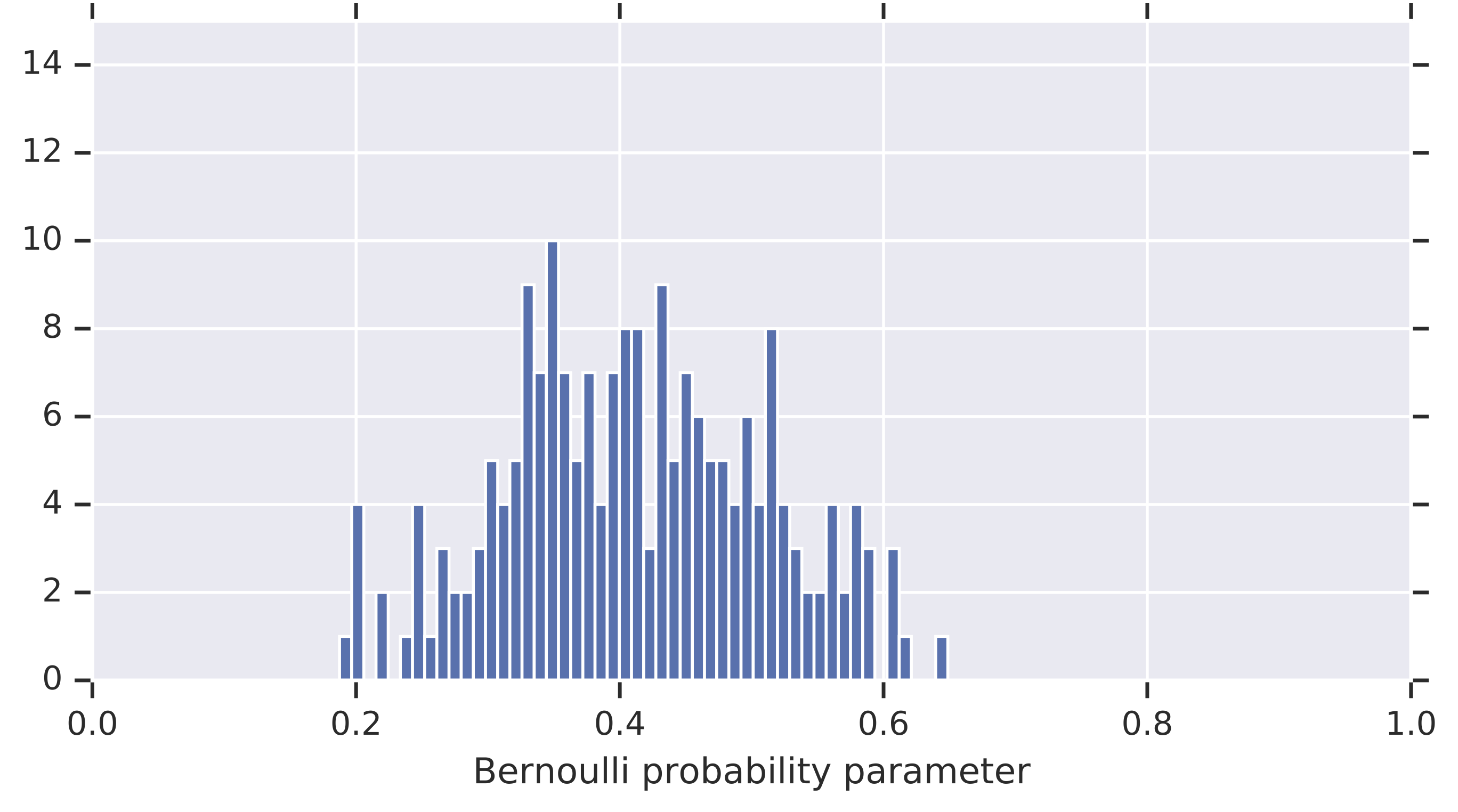}
\caption{A histogram of predictions from $M$ deterministic \gls*{rnn} models trained with different random seeds for a single \gls*{icu} patient's probability of mortality. 
As shown here, model uncertainty can cause high disagreement between individual models in an ensemble regarding the correct predictive distribution for a given patient.
This is not captured when using a single model or an average over an ensemble.}
\label{fig:pred_dist-deterministic}
\end{figure}

\subsection{Deep Ensembles}
Deep ensembles \citep{lakshminarayanan2017} is a method for quantifying model uncertainty.  In this approach, an ensemble of $M$ deterministic\footnote{We use the term ``deterministic" to refer to the usual setup in which we optimize the parameter values of our function directly, yielding a trained model with fixed parameter values at test time.} \glspl*{nn} is trained by varying only the random seed of an otherwise well-tuned set of hyperparameters.  Given this ensemble, a prediction $\boldsymbol{\lambda}^{(m)}$ can be made with each model $m$ for a given input $\mathbf{x}$, where (for a binary task) each prediction is the probability parameter for the Bernoulli distribution over the outcome.  The set of $M$ probabilistic predictions $\{\boldsymbol{\lambda}^{(1)}, \boldsymbol{\lambda}^{(2)}, \ldots, \boldsymbol{\lambda}^{(M)}\}$ for the same example can then be viewed as samples from the distribution $p(\boldsymbol{\lambda} | \mathbf{x})$ (equation \ref{eq:change-of-vars}),
where this distribution represents disagreement, or uncertainty due to model uncertainty.
In this work, we make use of deep ensembles of \glspl*{rnn} to model sequential patient data.

\subsection{Bayesian RNNs}
Bayesian \glspl*{rnn} \citep{fortunato2017} are \glspl*{rnn} with a prior distribution  $p(\mathbf{w})$ placed over the parameters $\mathbf{w}$ of the model.  This allows us to express model uncertainty as uncertainty over the true values for the parameters in the model, \textit{i.e.}, ``weight uncertainty" \citep{blundell2015}.  By introducing a distribution over all, or a subset, of the weights in the model, we can induce different functions, and thus different outcomes, through realizations of different weight values via draws from the posterior distributions over those weights.  This allows us to empirically capture model uncertainty in the predictive uncertainty distribution $p(\boldsymbol{\lambda} | \mathbf{x})$ by drawing $M$ samples from a single Bayesian \gls*{rnn} for a given example. In this work, we make use of various Bayesian \gls*{rnn} variants by placing priors on different subsets of the parameters.

 \begin{figure}
\centering
\includegraphics[width=\columnwidth]{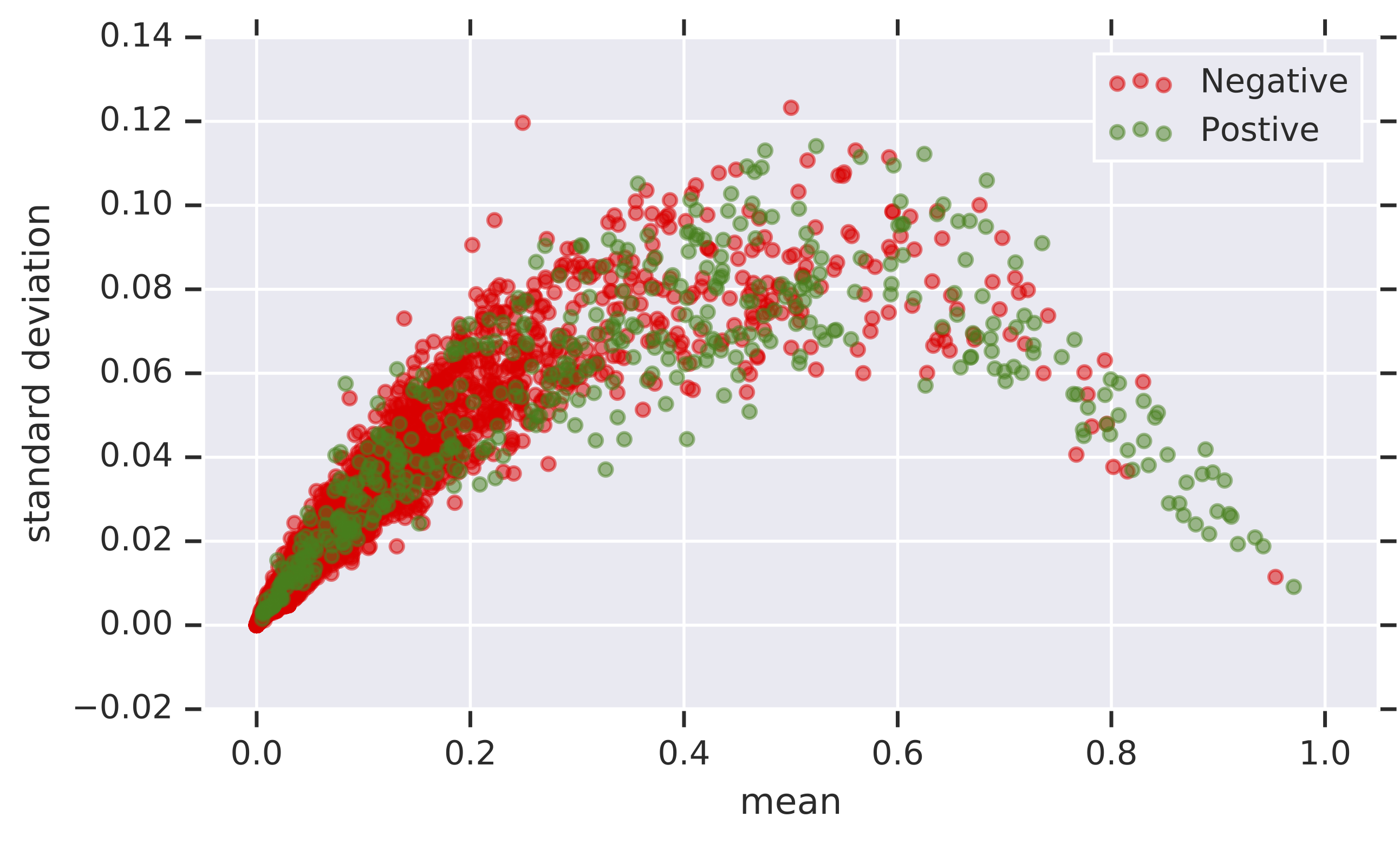}
\caption{A plot of the mean versus standard deviation of the predictive uncertainty distributions of the deterministic ensemble for positive and negative patients in the validation set.
We find that the standard deviations
do not form a simple linear relationship with the mean. For reference, we note that the variances of the distributions are generally lower than that of a Bernoulli distribution's variance curve.
}
\label{fig:pred_dist-mean-stddev-determininistic}
\end{figure}

\section{Medical Uncertainty}

\subsection{Clinical Tasks}
We demonstrate results on both binary and multiclass clinical tasks using multiple \gls*{ehr} datasets.  In terms of data, we use
\begin{enumerate}
    \item \gls*{mimic} \citep{johnson2016}, and
    \item \gls*{eicu} \citep{pollard2018eicu},
\end{enumerate}
both of which are publicly available \gls*{ehr} datasets.
\acrshort*{mimic} is collected from
46,520 patients admitted to \glspl*{icu} at Beth Israel Deaconess Medical Center, where 9,974 expired during the encounter (\textit{i.e.}, 1:4 ratio between positive and negative samples). The \acrshort*{eicu} dataset is collected from over 200,000 admissions to \glspl*{icu} across the United States.
In terms of tasks, for \acrshort*{mimic} we study
\begin{enumerate}
    \item binary in-patient mortality prediction, and
    \item multiclass diagnosis prediction at discharge.
\end{enumerate}
For the multiclass diagnosis prediction, we use the single-level \gls*{ccs} code system. For the \acrshort*{eicu} dataset, we study the binary in-patient mortality prediction task as well, allowing us to demonstrate that our findings generalize to additional datasets.

\subsection{Models}
Similar to \citet{rajkomar2018}, we train deep \glspl*{rnn} for our clinical tasks. Each of our models embeds and aggregates a patient's sequential features (\textit{e.g.} medications, lab measures, clinical notes) and global contextual features (\textit{e.g.} gender, age), feeds them to one or more \gls*{lstm} layers \citep{schmidhuber1997}, and follows that with hidden and output affine layers. 
More specifically, sequential embeddings are bagged into 1-day blocks, and fed into one or more \gls*{lstm} layers.  The final time-step output of the \gls*{lstm} layers is concatenated with the contextual embeddings and fed into a hidden dense layer, and the output of that layer is then fed into an output dense layer yielding the parameterization $\boldsymbol{\lambda}$ for a predictive distribution $p(y | \mathbf{x}, \mathbf{w})$.  A ReLU non-linearity is used between the hidden and output dense layers, and default initializers in \textrm{tf.keras.layers.*} are used for all deterministic layers.  More details on the training setup can be found in the Appendix and in the code\footnote{Code can be found at \url{https://github.com/Google-Health/records-research}.}

Existing deep learning approaches in predictive medicine focus on capturing data uncertainty, namely accurately predicting the
predictive distribution $p(y | \mathbf{x})$ of a patient outcome (\textit{i.e.}, how likely is the patient to expire?). This work, on the other hand, also focuses on addressing the model uncertainty aspect of deep learning, namely the distribution
over equally-likely predictive distributions (\textit{i.e.}, are there alternative predictive distributions, and if so, how diverse are the distributions?).

\begin{figure}
\centering
\includegraphics[width=\columnwidth]{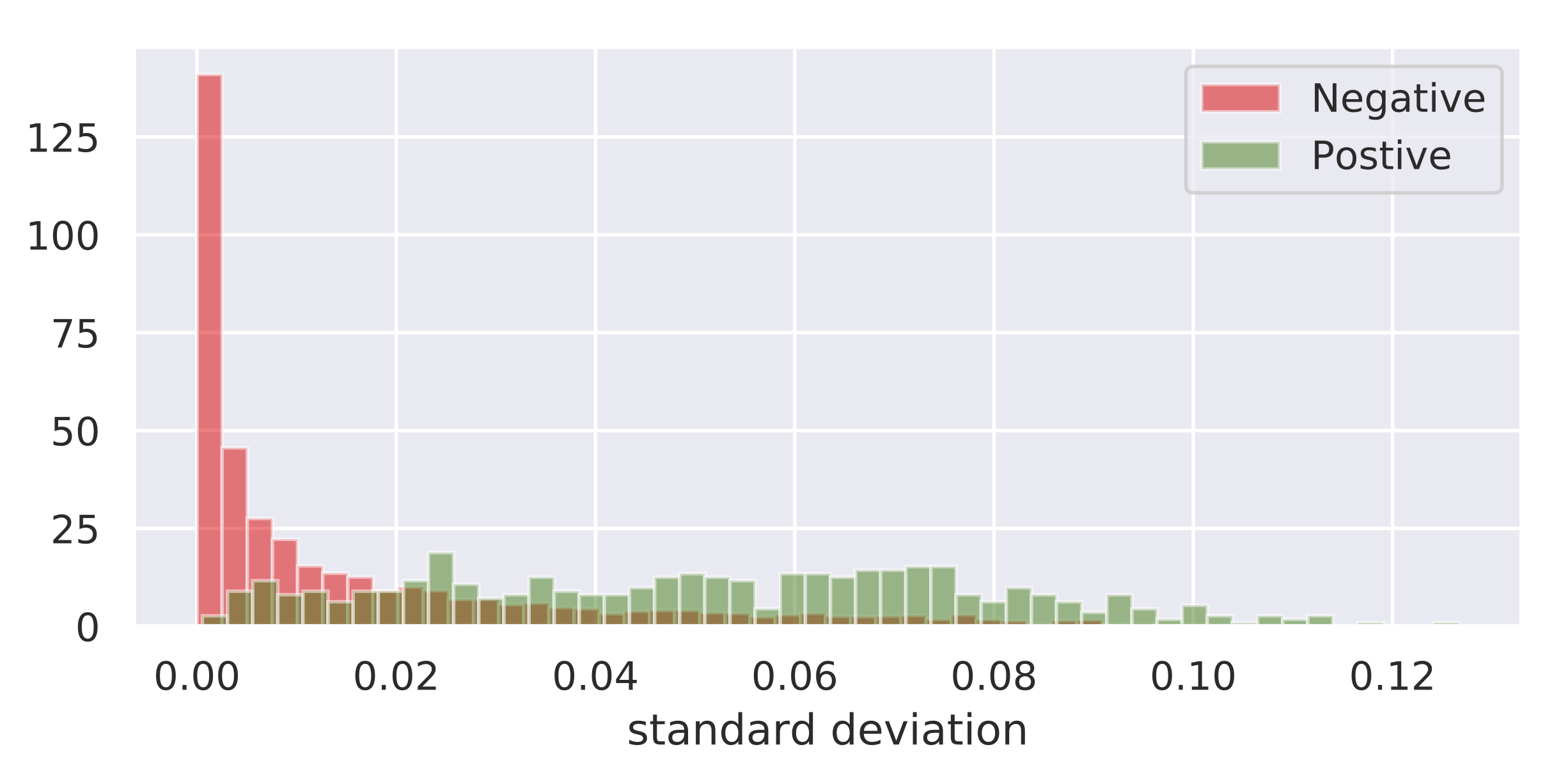}
\begin{flushright}\includegraphics[width=0.985\columnwidth]{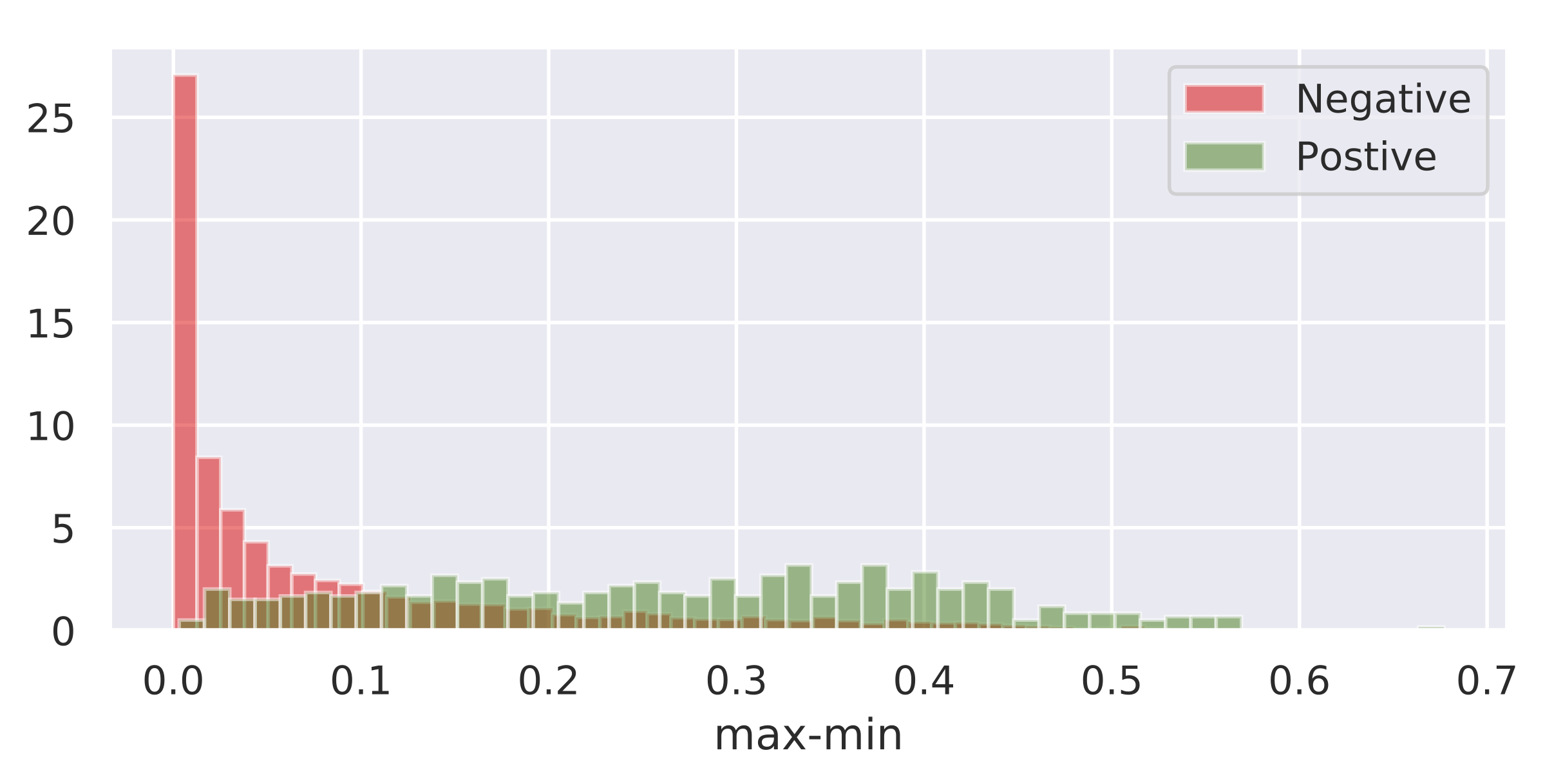}\end{flushright}
\caption{Top: A histogram of the standard deviations of the $p(\boldsymbol{\lambda} | \mathbf{x})$ distributions for all patients in the test set. Bottom: The same setup, but looking at differences between the maximum and minimum values of those $p(\boldsymbol{\lambda} | \mathbf{x})$ distributions.
Together, this shows that there is wide variability in predicted probabilities for some patients, and that negative patients have less variability on average.
}
\label{fig:pred-std-deterministic}
\end{figure}

\subsection{Choice of Uncertainty Methods}

To quantify model uncertainty for clinical tasks, we explore the use of deep \gls*{rnn} ensembles and various Bayesian \glspl*{rnn}.  For the deep ensembles approach, we optimize for the ideal hyperparameter values for our \gls*{rnn} model via black-box Bayesian optimization \citep{golovin2017}, and then train $M$ replicas of the best model.  Only the random seed differs between the replicas.  At prediction time, we make predictions with each of the $M$ models for each patient.  The full list of hyperparameters and the specific hyperparameter values for all models can be found in
Tables \ref{tab:hyp-search} and \ref{tab:hyp-values} in
the Appendix.

For the Bayesian \glspl*{rnn}, we train a single model, and then draw $M$ samples from it at prediction time. To train the Bayesian \gls*{rnn}, we take a variational inference approach by adapting our \gls*{rnn} to use factorized weight posteriors
\begin{equation*}
q(\mathbf{w} | \boldsymbol{\theta}) = \prod_i q(\mathbf{w}^{(i)} | \boldsymbol{\theta}^{(i)}), 
\end{equation*}
where each weight tensor $\mathbf{w}^{(i)}$ in the model is represented by a normal distribution with learnable mean and diagonal covariance parameters represented collectively as $\boldsymbol{\theta}^{(i)}$. Normal distributions with zero mean and tunable standard deviation are used as weight priors $p(\mathbf{w}^{(i)})$.  We train our models by minimizing the \gls*{kl} divergence
\begin{equation}
\begin{split}
\mathcal{L}(\boldsymbol{\theta}) &= \kl{q(\mathbf{w} | \boldsymbol{\theta})}{p(\mathbf{w} | \mathbf{y}, \mathbf{X})} \\
&\propto \kl{q(\mathbf{w} | \boldsymbol{\theta})}{p(\mathbf{w})} - \mathbb{E}_{q(\mathbf{w} | \boldsymbol{\theta})} \left[ \ln p(\mathbf{y} | \mathbf{X}, \mathbf{w}) \right]
\end{split}
\end{equation}
between the approximate weight posterior $q(\mathbf{w} | \boldsymbol{\theta})$ and the true, but unknown posterior $p(\mathbf{w} | \mathbf{y}, \mathbf{X})$. Overall, this equates to minimizing an expectation over the usual negative log likelihood term, $\mathbb{E}_{q(\mathbf{w} | \boldsymbol{\theta})} \left[ \ln p(\mathbf{y} | \mathbf{X}, \mathbf{w}) \right]$, plus a \gls*{kl} divergence regularization term.  To easily shift between the deterministic \gls*{rnn} and various Bayesian \gls*{rnn} models, we make use of the Bayesian Layers \citep{tran2018} abstractions.

\subsection{Optimal Decisions via Sensitivity Requirements}

The key desire in clinical practice is to make a decision based on the model's predicted probability and its associated uncertainty.  Given a set of potential outcomes $y_k \in \{1, \ldots, K\}$ for $K$ classes, a set of conditional probabilities $p(y_k | \mathbf{x})$ for the given outcomes, and the associated costs $L_{kj}$ for
predicting class $j$ when the true class is $k$, an optimal decision can be determined by minimizing the expected decision cost
\begin{equation}
\mathbb{E}[L] = \sum_k \sum_j \int_{\mathcal{R}_j} L_{kj} p(y_k | \mathbf{x})p(\mathbf{x}) \d{\mathbf{x}},
\end{equation}
with respect to $\mathcal{R}_j$, where $\mathcal{R}_j$ is the decision region for assigning example $\mathbf{x}$ to class $j$, and $p(\mathbf{x})$ is the density of $\mathbf{x}$ \citep{bishop2006}.

Designing elaborate decision cost functions for clinical applications is an interesting but difficult task, as it requires expert knowledge of the prediction target, cost-benefit analysis, and medical resource allocation. Fortunately, we can use a clinically relevant alternative, which is the \textit{sensitivity requirement}.  Often in clinical research, certain sensitivity (\textit{i.e.}, recall) levels
are desirable when making predictions in order for a model to be clinically relevant \citep{stiell2001,stiell2005comparison,smits2005,reynolds2013tunisian,dusenberry2017}.
The goal in such cases is to maximize the precision while still maintaining the
desired sensitivity level.
Viewed as a decision cost function, the cost is infinite if the recall is below the target level, and is otherwise minimized as the precision is increased, where the optimized parameter is a global probability threshold $t^{(m)}$ for a given model $m$.

For each of the $M$ models in our ensemble, we can optimize the sensitivity-based decision cost function and make optimal decisions for all examples.  Thus, for each example, there will be a set of $M$ optimal decisions, which can be represented as a distribution.
That is, from this viewpoint, the optimal decision $d$ for an example $\mathbf{x}$ can be represented as a random variable
\begin{equation}
    d \sim p(d | \mathbf{x}),
\end{equation}
which, for a binary task, can be approximated as
\begin{equation}
\begin{split}
\phi &= \frac{1}{M} \sum_{m=1}^M \mathds{1}(\boldsymbol{\lambda}^{(m)} \ge t^{(m)})) \\
d &\sim \operatorname{Bernoulli}(\phi),
\end{split}
\end{equation}
where $\boldsymbol{\lambda}^{(m)} \ge t^{(m)}$ is the decision function for model $m$, and $\phi$ is the percentage of model agreement.

This simply represents the propagation of uncertainty over functions into uncertainty over predictive distributions (equation \ref{eq:change-of-vars}), and, in turn, into uncertainty over optimal decisions. That is, different equally-likely functions could yield different values for $\boldsymbol{\lambda}$ and thus different predictive distributions $p(y | \mathbf{x}, \mathbf{w})$ for a given example, which could lead to different optimal decisions for that example. In the same way that we could represent a set of functions as a distribution over functions, we could represent a set of predictive distributions as a distribution over predictive distributions, and we could represent a set of optimal decisions as a distribution over optimal decisions. As stated previously, the variance of these distributions represents \textit{disagreement}, \textit{i.e.}, uncertainty due to model uncertainty.

 \section{Experiments}

We perform four sets of experiments. First, in order to demonstrate the importance of quantifying uncertainty in predictive medicine, we examine individual models in the \gls*{rnn} ensemble in terms of predictive metrics, calibration, uncertainty distributions, and decision-making.
Second, we examine multiple variants of Bayesian \glspl*{rnn} to understand where uncertainty in the model matters most, comparing them with their deterministic ensemble counterpart.
Third, we use the deterministic \gls*{rnn} ensemble to examine uncertainty across different patient subgroups.
Finally, we analyze the Bayesian \gls*{rnn} with embedding distributions to examine uncertainty across individual features.

\subsection{When Do We Observe Uncertainty?}

\paragraph{Clinical Metrics}
For our clinical tasks, we first measure the dataset-level metrics:
\begin{itemize}
    \item \gls*{aucpr} (\textit{binary tasks}),
    \item \gls*{aucroc} (\textit{binary tasks}),
    \item top-5 recall (\textit{multiclass tasks}),
    \item top-5 precision (\textit{multiclass tasks}),
    \item top-5 F1 (\textit{multiclass tasks}),
    \item held-out negative log-likelihood (\textit{all tasks}),
    \item \gls*{ece} (\textit{all tasks}) \citep{naeini2015}, and
    \item \gls*{ace} (\textit{all tasks}) \citep{nixon2019}.
\end{itemize} \autoref{tab:ensemble-metrics} shows the performance on the \gls*{mimic} binary mortality and multiclass \gls*{ccs} multiclass tasks averaged over individual models in our deterministic \gls*{rnn} ensemble, with the standard deviation over models in the parentheses. Interestingly, individual models are overall well-calibrated and \textit{nearly equivalent in terms of likelihood and metric performance}. If we were to choose only one model in practice based on the dataset-level metrics, it is highly likely any of the models in the ensemble could be selected. Importantly, if we only used a single model, we would lose the model uncertainty information (as noted in Section \ref{sec:model-uncertainty}).

\begin{table}
\caption{Dataset-level metrics for the \gls*{mimic} binary mortality and multiclass \acrshort*{ccs} prediction tasks across $M = 200$ models in the deterministic \gls*{rnn} ensemble.  Metrics are computed for each model within the ensemble, and means and standard deviations across models are reported. Individual models are nearly identical in terms of dataset-level performance across both tasks, but selecting a single model would remove the model uncertainty information such as that visualized in \autoref{fig:pred_dist-deterministic}.
}
\label{tab:ensemble-metrics}
\begin{center}
\begin{small}
\begin{sc}
\resizebox{\columnwidth}{!}{
\begin{tabular}{c l c c}
\toprule
& Metric & Validation & Test  \\
\midrule
\multirow{5}{*}{\rotatebox[origin=c]{90}{Mortality}}
& \acrshort*{aucpr} $\uparrow$ & $0.4496$ ($0.0025$) & $0.3886$ ($0.0059$) \\
& \acrshort*{aucroc} $\uparrow$ & $0.8753$ ($0.0019$) & $0.8623$ ($0.0031$) \\
& Neg. Log-likelihood $\downarrow$ & $0.2037$ ($0.0030$) & $0.2088$ ($0.0038$) \\
& \acrshort*{ece} $\downarrow$ & $0.0176$ ($0.0040$) & $0.0162$ ($0.0043$) \\
& \acrshort*{ace} $\downarrow$ & $0.0210$ ($0.0042$) & $0.0233$ ($0.0057$) \\
\midrule
\multirow{6}{*}{\rotatebox[origin=c]{90}{CCS Diagnosis}}
& Top-5 recall $\uparrow$ & $0.7126$ ($0.0071$) & $0.7090$ ($0.0088$) \\
& Top-5 precision $\uparrow$ & $0.1425$ ($0.0014$) & $0.1418$ ($0.0018$) \\
& Top-5 F1 $\uparrow$ & $0.2375$ ($0.0024$) & $0.2363$ ($0.0029$) \\
& Neg. Log-likelihood $\downarrow$ & $2.2738$ ($0.0330$) & $2.3338$ ($0.0434$) \\
& \acrshort*{ece} $\downarrow$ & $0.0446$ ($0.0072$) & $0.0499$ ($0.0082$) \\
& \acrshort*{ace} $\downarrow$ & $4.219\mathrm{e}{-3}$ ($7.31\mathrm{e}{-8}$) & $4.219\mathrm{e}{-3}$ ($7.61\mathrm{e}{-8}$) \\
\bottomrule
\end{tabular}
}
\end{sc}
\end{small}
\end{center}
\end{table}

\begin{figure*}
\begin{center}
\includegraphics[width=\textwidth]{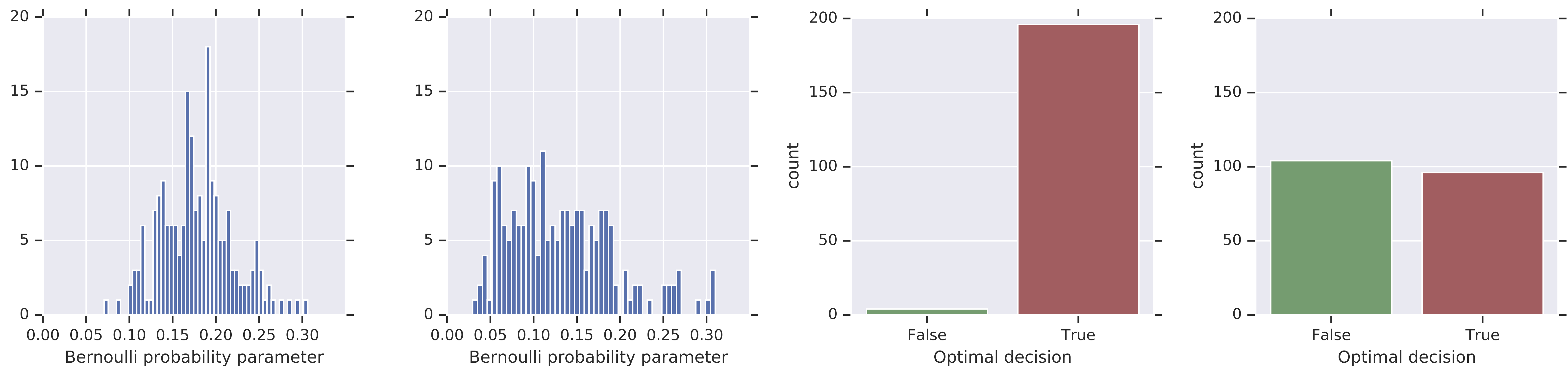}
\end{center}
\caption{
\textit{Left Two:}
Histograms representing the different mortality predictive distributions produced by the deterministic ensemble
for two patients in the validation set. 
\,\textit{Right Two:}
The corresponding optimal decision distributions, with ``True'' corresponding to a prediction of mortality and ``False'' corresponding to the opposite.
For one patient, the ensemble members are in agreement about the optimal decision, while for the other patient there is high disagreement due to model uncertainty.}
\label{fig:dist_comparisons-deterministic}
\end{figure*}

\begin{table*}[t]
\caption{Metrics for marginalized predictions on the \gls*{mimic} and \gls*{eicu} mortality tasks given $M = 200$ models in the deterministic \gls*{rnn} ensemble, and $M = 200$ samples from each of the Bayesian \gls*{rnn} models.  95\% confidence intervals are computed via validation and test set bootstrapping with 1000 bootstrap sets.
}
\label{tab:marginalized-metrics}
\begin{center}
\begin{small}
\begin{sc}
\resizebox{\textwidth}{!}{
\begin{tabular}{l l | c c c | c c c}
\toprule
& Model
& \begin{tabular}{@{}c@{}} Val.\\\,\,\acrshort*{aucpr} $\uparrow$\end{tabular}
& \begin{tabular}{@{}c@{}} Val.\\\,\,\acrshort*{aucroc} $\uparrow$\end{tabular}
& \begin{tabular}{@{}c@{}} Val.\\\,\,\,NLL $\downarrow$\end{tabular}
& \begin{tabular}{@{}c@{}} Test\\\,\,\,\acrshort*{aucpr} $\uparrow$\end{tabular}
& \begin{tabular}{@{}c@{}} Test\\\,\,\,\acrshort*{aucroc} $\uparrow$\end{tabular}
& \begin{tabular}{@{}c@{}} Test\\\,\,\,NLL $\downarrow$\end{tabular}
\\
\midrule
\multirow{6}{*}{\rotatebox[origin=c]{90}{\gls*{mimic}}}
& \begin{tabular}{@{}l@{}}Deterministic Ensemble\end{tabular}
& \begin{tabular}{@{}c@{}} $0.4564$ $(\pm 1\mathrm{e}{-3})$ \end{tabular}
& \begin{tabular}{@{}c@{}} $0.8774$ $(\pm 5\mathrm{e}{-4})$ \end{tabular}
& \begin{tabular}{@{}c@{}} $0.1999$ $(\pm 5\mathrm{e}{-4})$ \end{tabular}
& \begin{tabular}{@{}c@{}} $0.3921$ $(\pm 1\mathrm{e}{-3})$ \end{tabular}
& \begin{tabular}{@{}c@{}} $0.8643$ $(\pm 5\mathrm{e}{-4})$ \end{tabular}
& \begin{tabular}{@{}c@{}} $0.2051$ $(\pm 5\mathrm{e}{-4})$ \end{tabular}
\\

&\begin{tabular}{@{}l@{}}Bayesian Embeddings\end{tabular}
& \begin{tabular}{@{}c@{}} $0.4580$ $(\pm 1\mathrm{e}{-3})$ \end{tabular}
& \begin{tabular}{@{}c@{}} $0.8776$ $(\pm 4\mathrm{e}{-4})$ \end{tabular}
& \begin{tabular}{@{}c@{}} $0.2002$ $(\pm 4\mathrm{e}{-4})$ \end{tabular}
& \begin{tabular}{@{}c@{}} $0.3977$ $(\pm 2\mathrm{e}{-3})$ \end{tabular}
& \begin{tabular}{@{}c@{}} $0.8612$ $(\pm 5\mathrm{e}{-4})$ \end{tabular}
& \begin{tabular}{@{}c@{}} $0.2059$ $(\pm 4\mathrm{e}{-4})$ \end{tabular}
\\

&\begin{tabular}{@{}l@{}}Bayesian Output\end{tabular}
& \begin{tabular}{@{}c@{}} $0.4382$ $(\pm 2\mathrm{e}{-3})$ \end{tabular}
& \begin{tabular}{@{}c@{}} $0.8714$ $(\pm 5\mathrm{e}{-4})$ \end{tabular}
& \begin{tabular}{@{}c@{}} $0.2189$ $(\pm 6\mathrm{e}{-4})$ \end{tabular}
& \begin{tabular}{@{}c@{}} $0.3702$ $(\pm 1\mathrm{e}{-3})$ \end{tabular}
& \begin{tabular}{@{}c@{}} $0.8572$ $(\pm 5\mathrm{e}{-4})$ \end{tabular}
& \begin{tabular}{@{}c@{}} $0.2246$ $(\pm 6\mathrm{e}{-4})$ \end{tabular}
\\

&\begin{tabular}{@{}l@{}}Bayesian Hidden+Output\end{tabular}
& \begin{tabular}{@{}c@{}} $0.4492$ $(\pm 1\mathrm{e}{-3})$ \end{tabular}
& \begin{tabular}{@{}c@{}} $0.8751$ $(\pm 5\mathrm{e}{-4})$ \end{tabular}
& \begin{tabular}{@{}c@{}} $0.2045$ $(\pm 5\mathrm{e}{-4})$ \end{tabular}
& \begin{tabular}{@{}c@{}} $0.3893$ $(\pm 1\mathrm{e}{-3})$ \end{tabular}
& \begin{tabular}{@{}c@{}} $0.8607$ $(\pm 5\mathrm{e}{-4})$ \end{tabular}
& \begin{tabular}{@{}c@{}} $0.2102$ $(\pm 5\mathrm{e}{-4})$ \end{tabular}
\\

&\begin{tabular}{@{}l@{}}Bayesian RNN+Hidden+Output\end{tabular}
& \begin{tabular}{@{}c@{}} $0.4396$ $(\pm 2\mathrm{e}{-3})$ \end{tabular}
& \begin{tabular}{@{}c@{}} $0.8673$ $(\pm 5\mathrm{e}{-4})$ \end{tabular}
& \begin{tabular}{@{}c@{}} $0.2109$ $(\pm 5\mathrm{e}{-4})$ \end{tabular}
& \begin{tabular}{@{}c@{}} $0.3860$ $(\pm 2\mathrm{e}{-3})$ \end{tabular}
& \begin{tabular}{@{}c@{}} $0.8542$ $(\pm 5\mathrm{e}{-4})$ \end{tabular}
& \begin{tabular}{@{}c@{}} $0.2146$ $(\pm 5\mathrm{e}{-4})$ \end{tabular}
\\

&\begin{tabular}{@{}l@{}}Fully Bayesian\end{tabular}
& \begin{tabular}{@{}c@{}} $0.4354$ $(\pm 2\mathrm{e}{-3})$ \end{tabular}
& \begin{tabular}{@{}c@{}} $0.8692$ $(\pm 5\mathrm{e}{-4})$ \end{tabular}
& \begin{tabular}{@{}c@{}} $0.2068$ $(\pm 5\mathrm{e}{-4})$ \end{tabular}
& \begin{tabular}{@{}c@{}} $0.3829$ $(\pm 1\mathrm{e}{-3})$ \end{tabular}
& \begin{tabular}{@{}c@{}} $0.8552$ $(\pm 5\mathrm{e}{-4})$ \end{tabular}
& \begin{tabular}{@{}c@{}} $0.2103$ $(\pm 5\mathrm{e}{-4})$ \end{tabular}
\\

\midrule
\multirow{6}{*}{\rotatebox[origin=c]{90}{\gls*{eicu}}}
& \begin{tabular}{@{}l@{}}Deterministic Ensemble\end{tabular}
& \begin{tabular}{@{}c@{}} $0.1951$ $(\pm 1\mathrm{e}{-3})$ \end{tabular}
& \begin{tabular}{@{}c@{}} $0.7882$ $(\pm 7\mathrm{e}{-4})$ \end{tabular}
& \begin{tabular}{@{}c@{}} $0.1435$ $(\pm 3\mathrm{e}{-4})$ \end{tabular}
& \begin{tabular}{@{}c@{}} $0.2196$ $(\pm 1\mathrm{e}{-3})$ \end{tabular}
& \begin{tabular}{@{}c@{}} $0.7868$ $(\pm 6\mathrm{e}{-4})$ \end{tabular}
& \begin{tabular}{@{}c@{}} $0.2435$ $(\pm 5\mathrm{e}{-4})$ \end{tabular}
\\

&\begin{tabular}{@{}l@{}}Bayesian Embeddings\end{tabular}
& \begin{tabular}{@{}c@{}} $0.1996$ $(\pm 1\mathrm{e}{-3})$ \end{tabular}
& \begin{tabular}{@{}c@{}} $0.7807$ $(\pm 1\mathrm{e}{-4})$ \end{tabular}
& \begin{tabular}{@{}c@{}} $0.1455$ $(\pm 4\mathrm{e}{-4})$ \end{tabular}
& \begin{tabular}{@{}c@{}} $0.2244$ $(\pm 1\mathrm{e}{-3})$ \end{tabular}
& \begin{tabular}{@{}c@{}} $0.7733$ $(\pm 7\mathrm{e}{-4})$ \end{tabular}
& \begin{tabular}{@{}c@{}} $0.1620$ $(\pm 4\mathrm{e}{-4})$ \end{tabular}
\\

&\begin{tabular}{@{}l@{}}Bayesian Output\end{tabular}
& \begin{tabular}{@{}c@{}} $0.1738$ $(\pm 1\mathrm{e}{-3})$ \end{tabular}
& \begin{tabular}{@{}c@{}} $0.7677$ $(\pm 7\mathrm{e}{-4})$ \end{tabular}
& \begin{tabular}{@{}c@{}} $0.1664$ $(\pm 3\mathrm{e}{-4})$ \end{tabular}
& \begin{tabular}{@{}c@{}} $0.1942$ $(\pm 1\mathrm{e}{-3})$ \end{tabular}
& \begin{tabular}{@{}c@{}} $0.7580$ $(\pm 7\mathrm{e}{-4})$ \end{tabular}
& \begin{tabular}{@{}c@{}} $0.1810$ $(\pm 4\mathrm{e}{-4})$ \end{tabular}
\\

&\begin{tabular}{@{}l@{}}Bayesian Hidden+Output\end{tabular}
& \begin{tabular}{@{}c@{}} $0.1712$ $(\pm 1\mathrm{e}{-3})$ \end{tabular}
& \begin{tabular}{@{}c@{}} $0.7801$ $(\pm 7\mathrm{e}{-4})$ \end{tabular}
& \begin{tabular}{@{}c@{}} $0.1619$ $(\pm 3\mathrm{e}{-4})$ \end{tabular}
& \begin{tabular}{@{}c@{}} $0.2140$ $(\pm 1\mathrm{e}{-3})$ \end{tabular}
& \begin{tabular}{@{}c@{}} $0.7817$ $(\pm 6\mathrm{e}{-4})$ \end{tabular}
& \begin{tabular}{@{}c@{}} $0.1713$ $(\pm 3\mathrm{e}{-4})$ \end{tabular}
\\

&\begin{tabular}{@{}l@{}}Bayesian RNN+Hidden+Output\end{tabular}
& \begin{tabular}{@{}c@{}} $0.1675$ $(\pm 1\mathrm{e}{-3})$ \end{tabular}
& \begin{tabular}{@{}c@{}} $0.7791$ $(\pm 7\mathrm{e}{-4})$ \end{tabular}
& \begin{tabular}{@{}c@{}} $0.1477$ $(\pm 3\mathrm{e}{-4})$ \end{tabular}
& \begin{tabular}{@{}c@{}} $0.2147$ $(\pm 1\mathrm{e}{-3})$ \end{tabular}
& \begin{tabular}{@{}c@{}} $0.7809$ $(\pm 7\mathrm{e}{-4})$ \end{tabular}
& \begin{tabular}{@{}c@{}} $0.1583$ $(\pm 3\mathrm{e}{-4})$ \end{tabular}
\\

&\begin{tabular}{@{}l@{}}Fully Bayesian\end{tabular}
& \begin{tabular}{@{}c@{}} $0.2004$ $(\pm 1\mathrm{e}{-3})$ \end{tabular}
& \begin{tabular}{@{}c@{}} $0.7910$ $(\pm 7\mathrm{e}{-4})$ \end{tabular}
& \begin{tabular}{@{}c@{}} $0.1377$ $(\pm 3\mathrm{e}{-4})$ \end{tabular}
& \begin{tabular}{@{}c@{}} $0.2280$ $(\pm 1\mathrm{e}{-3})$ \end{tabular}
& \begin{tabular}{@{}c@{}} $0.7818$ $(\pm 7\mathrm{e}{-4})$ \end{tabular}
& \begin{tabular}{@{}c@{}} $0.1541$ $(\pm 4\mathrm{e}{-4})$ \end{tabular}
\\

\bottomrule
\end{tabular}
}
\end{sc}
\end{small}
\end{center}
\end{table*} 

\paragraph{Predictive Uncertainty Distributions \& Statistics}
Knowing that the models in our ensemble are well-calibrated and effectively equivalent in terms of performance, we turn to making predictions for individual examples.  \autoref{fig:pred_dist-deterministic} visualizes the predictive uncertainty distribution for a single patient on the mortality task using the deterministic \gls*{rnn} ensemble.  We find that there is a wide variability in predicted Bernoulli probabilities for some patients (with spreads as high as $57.5\%$).  As noted in Section \ref{sec:model-uncertainty}, this variability represents our uncertainty associated with determining the correct predictive distribution $p(y | \mathbf{x}, \mathbf{w})$ for the given patient. Marginalizing over this uncertainty with respect to $\mathbf{w}$ will yield the current best estimate for $p(y | \mathbf{x})$, but the estimate could be improved through the acquisition of more training examples similar to the current patient. Ignoring the variance $\operatorname{Var}[\boldsymbol{\lambda} | \mathbf{x}]$ through the use of either a single model or an average over models without also conveying the original variance is likely detrimental since 
it is not possible to distinguish between data uncertainty and model uncertainty from that marginalized distribution $p(y | \mathbf{x})$ alone, 
and thus it prevents a physician from being able to understand when a model is uncertain about the prediction it is making.

\autoref{fig:pred_dist-mean-stddev-determininistic} visualizes the means versus standard deviations of the predictive uncertainty distributions for the deterministic ensemble on all validation set examples.  In contrast to the variance of a Bernoulli distribution, which is a simple function of the mean, we find that the standard deviations are patient-specific, and thus cannot be determined \textit{a priori}. In \autoref{fig:pred-std-deterministic}, we plot the standard deviations and differences between the maximum and minimum predicted probability values for each patient's predictive uncertainty distribution, $p(\boldsymbol{\lambda} | \mathbf{x})$.  We find that there is wide variability in predicted probabilities for some patients, and that negative patients have less variability on average.

\begin{figure*}[ht]
\begin{center}
\includegraphics[width=\textwidth]{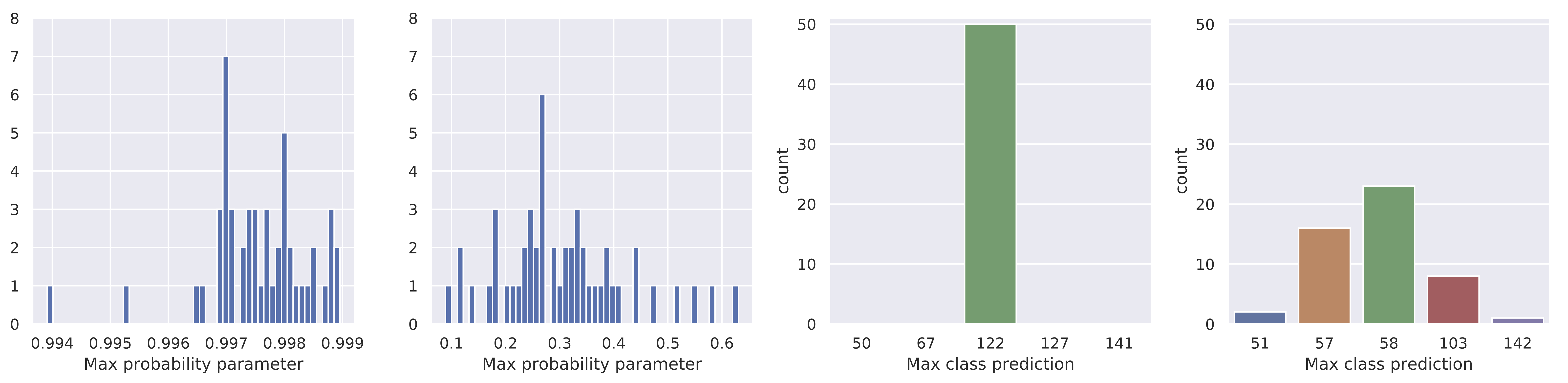}
\end{center}
\caption{\textit{Left two:} A set of distributions for the maximum predicted probability from our deterministic ensemble for two patients in the validation dataset on the multiclass CCS diagnosis code task.  Note the difference in x-axis scales. \,\textit{Right two:} The corresponding distributions of classes associated with the max probabilities.  Similar to the mortality task, for one patient,
the ensemble is relatively certain about the predicted class,
while for the other patient,
there is a larger amount of disagreement.
}
\label{fig:dist_comparisons-deterministic-ccs}
\end{figure*}

\paragraph{Optimal Decision Distributions \& Statistics}
In practice, model uncertainty is important insofar as it can affect the model's decisions. To test this, we optimize the sensitivity-based (\textit{i.e.}, recall-based) decision cost function with respect to the probability threshold for each model in our \gls*{rnn} ensemble separately to achieve a recall of $70\%$, and then make optimal decisions for each example with each of the $M$ models.
\autoref{fig:dist_comparisons-deterministic} visualizes how model uncertainty in probability space is realized in optimal decision space for two patients in the mortality task.
We see that the model uncertainty does indeed extend into the optimal decision space, leading to a set of optimal decisions for a given patient that can be represented as a distribution over the optimal decision. Furthermore, the decision distribution's variance can be quite high, and knowing when this is the case is important in order to avoid the cost of any incorrect decisions made by the system due to lack of precise knowledge about the correct predictive distribution $p(y | \mathbf{x})$ (\textit{i.e.}, the correct level of data uncertainty).

\autoref{fig:dist_comparisons-deterministic-ccs} examines the distribution of maximum predicted probabilities over the \gls*{ccs} classes, along with the distribution of predicted classes associated with the maximum probabilities. Similar to the binary mortality task, this demonstrates the presence of disagreement due to model uncertainty in the multiclass clinical setting.

\begin{figure*}[ht]
\begin{center}
\includegraphics[width=\textwidth]{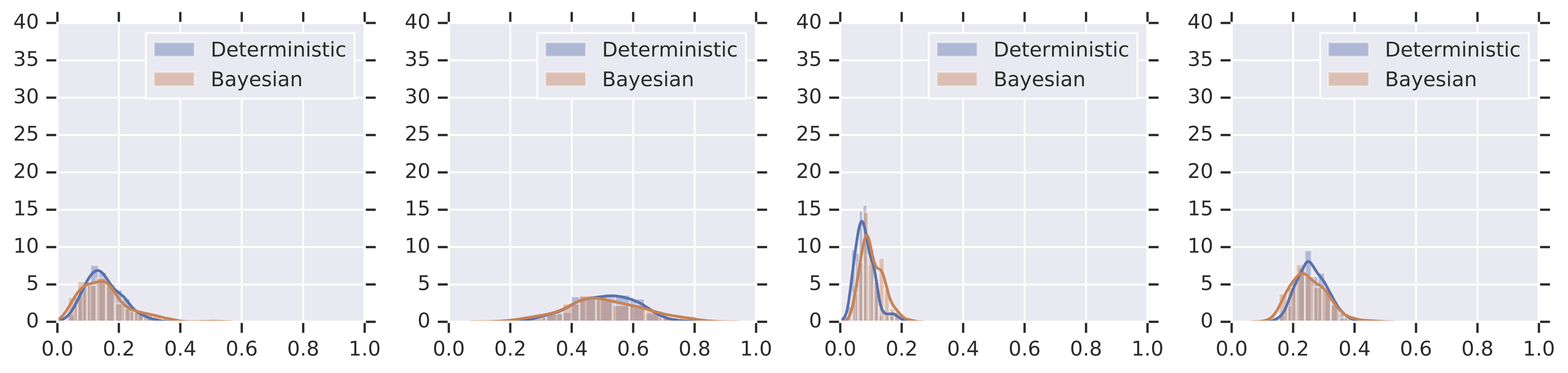}
\end{center}
\caption{Predictive uncertainty distributions of both the \gls*{rnn} with Bayesian embeddings and the deterministic \gls*{rnn} ensemble for individual patients. We find that the Bayesian model is qualitatively able to capture model uncertainty that aligns with that of the ensemble, while using a considerably smaller number of parameters.}
\label{fig:bayesian_vs_deterministic}
\end{figure*}

\subsection{Comparison: Variants of Bayesian RNNs and Deterministic RNN Ensembles}

A natural question in practice when employing the Bayesian approach is: which part of the model should capture model uncertainty?
To answer this question, we study Bayesian \glspl*{rnn} under a variety of priors:
\begin{itemize}
    \item \textbf{Bayesian Embeddings} A Bayesian \gls*{rnn} in which the embedding parameters are stochastic, and all other parameters are deterministic.
    \vskip 0.05in
    \item \textbf{Bayesian Output} A Bayesian \gls*{rnn} in which the output layer parameters are stochastic, and all other parameters are deterministic.
    \vskip 0.05in
    \item \textbf{Bayesian Hidden+Output} A Bayesian \gls*{rnn} in which the hidden and output layer parameters are stochastic, and all other parameters are deterministic.
    \vskip 0.05in
    \item \textbf{Bayesian RNN+Hidden+Output} A Bayesian \gls*{rnn} in which the \gls*{lstm}, hidden, and output layer parameters are stochastic, and all other parameters are deterministic.
    \vskip 0.05in
    \item \textbf{Fully Bayesian} A Bayesian \gls*{rnn} in which all parameters are stochastic.
\end{itemize}
\vspace{1em}

\autoref{tab:marginalized-metrics} displays \gls*{aucpr}, \gls*{aucroc}, and \gls*{nll} metrics over marginalized predictions for each of the Bayesian \gls*{rnn} models and the deterministic \gls*{rnn} ensemble on the \gls*{mimic} and \gls*{eicu} mortality tasks.
We find that the Bayesian Embeddings \gls*{rnn} model outperforms all other Bayesian variants and slightly outperforms the deterministic \gls*{rnn} ensemble in terms of \gls*{aucpr} for \gls*{mimic}, and that the fully-Bayesian \gls*{rnn} outperforms the other models on the \gls*{eicu} dataset.
Additionally, all of the Bayesian variants are either comparable or outperform the deterministic ensemble in terms of held-out \gls*{nll} on both datasets.

\begin{figure*}[ht]
\begin{center}
\includegraphics[width=\textwidth]{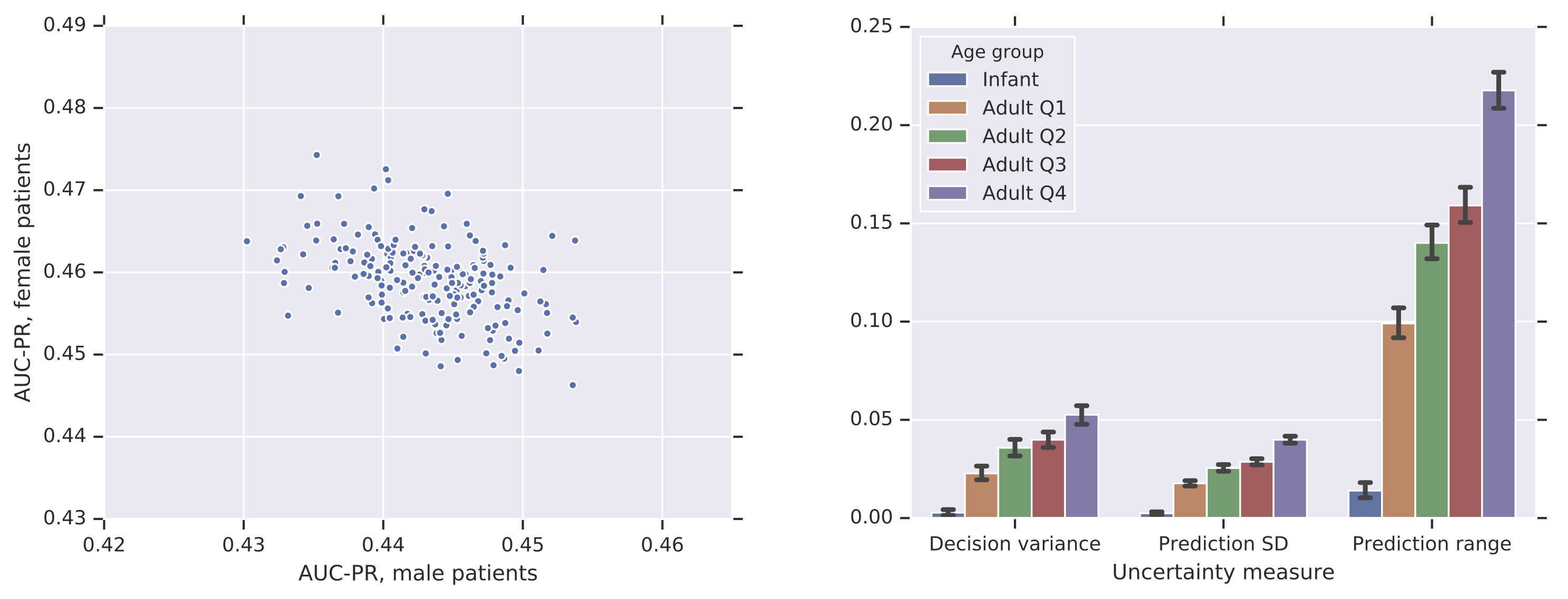}
\end{center}
\caption{\textbf{Left:} Model performance comparison on male vs. female patients. Each point represents stratified AUC-PR for a single model from the deterministic ensemble. Correlation coefficient $r = -0.442$. \textbf{Right:} Summary of uncertainty measures within each age subgroup, using the Bayesian Embeddings RNN. On all measures, uncertainty increases monotonically with age. This corresponds to an increase in mortality rate with age, as positive cases are more uncertain on average.}
\label{fig:subgroup_comparison}
\vspace{1em}
\end{figure*}

\autoref{fig:bayesian_vs_deterministic} visualizes the predictive distributions of both the Bayesian \gls*{rnn} with Bayesian embeddings, and the deterministic \gls*{rnn} ensemble for four individual patients on the \gls*{mimic} mortality task. The aim is to determine whether the two models are capturing the same distribution over functions insofar as they each produce the same distribution $p(\boldsymbol{\lambda} | \mathbf{x})$ for a given patient $\mathbf{x}$.  We find that the Bayesian model qualitatively captures model uncertainty that aligns with that of the deterministic ensemble. Overall, the Bayesian Embeddings \gls*{rnn}, compared to the deterministic \gls*{rnn} ensemble, demonstrates slightly improved predictive performance and qualitatively similar model uncertainty.

Our Bayesian models achieve strong performance while only requiring training of a single model ($7.22$ million parameters in the \gls*{mimic} Bayesian Embeddings \gls*{rnn}), versus $M$ models in the deterministic \gls*{rnn} ensemble ($M \times 6.16$ million parameters), as well as only requiring a single model at prediction time. In the deterministic ensemble case, we must choose the number of models $M$ \textit{a priori}, where $M$ affects the level of detail we can expect to obtain in our predictive distributions.  With a Bayesian model, we can choose the number of samples $M$ to draw at prediction time, dynamically adjusting it as we see fit. With considerably less computational resources required, using Bayesian \glspl*{rnn} can be a more efficient approach, making it an attractive choice for deployment in clinical practice.

\subsection{Patient Subgroup Analysis}

We next turn to an exploration of the effects of model uncertainty across patient subgroups.  We split validation set encounters into subgroups by demographic characteristics, namely patient gender (3089 male vs. 2548 female) and age (adults divided into quartiles of 1216, with a separate fifth group of 773 neonates). For this analysis, we focus on the deterministic \gls*{rnn} ensemble described in Section 4.1, as the Bayesian models sample $M = 200$ weights for each prediction separately rather than globally for repeated usage across the complete validation set. For each model in the ensemble, we compute validation set performance metrics separately over each subgroup and then compute the correlation between these metrics over all models in the ensemble to evaluate whether the ensemble models tend to specialize to one or more subgroups at the cost of performance on others. We find some evidence of this phenomenon: for example, AUC-PR for male patients is negatively correlated with AUC-PR for female patients (Pearson's $r = -0.442$, see \autoref{fig:subgroup_comparison}), and AUC-PR for the oldest quartile of adult patients is somewhat negatively correlated with AUC-PR for other adults or for neonates (Pearson's $r$ between $-0.18$ and $-0.37$).

We also compare uncertainty metrics across subgroups, including standard deviation and range of the predictive uncertainty distributions, and variance of the optimal decision distributions for patients in each subgroup. For this analysis, we examine both the deterministic \gls*{rnn} ensemble and the best Bayesian model, the \gls*{rnn} using Bayesian embeddings. In both cases, we find that all metrics are correlated with subgroup label prevalence: both uncertainty and mortality rate increase monotonically across age groups (\autoref{fig:subgroup_comparison}), and both are slightly higher in women than in men. These findings imply that random model variation during training may actually cause unintentional harm to certain patient populations, which may not be reflected in aggregate performance.

\subsection{Embedding Uncertainty Analysis}

\begin{table}
\caption{Top and bottom 10 words in free-text clinical notes with the highest and lowest model uncertainty based on the entropy of each word's associated Bayesian embedding distribution, along with total counts in the training set.}
\label{tab:notes-entropy-freq}
\begin{center}
\begin{scriptsize}
\begin{sc}
\resizebox{0.85\columnwidth}{!}{

\begin{tabular}{l c c}
\toprule
\multicolumn{3}{c}{\textbf{Lowest Uncertainty}}\\
Word & Entropy & Count \\ 
\midrule
the     & -82.5444 & 41803 \\
and     & -80.6054 & 42812 \\
of      & -80.2735 & 43191 \\
no      & -79.8993 & 43420 \\
tracing & -78.5987 & 32181 \\
is      & -78.5552 & 42560 \\
to      & -77.6408 & 42365 \\
for     & -76.8005 & 42972 \\
with    & -75.3513 & 42819 \\
in      & -72.8005 & 42144 \\

\midrule

\multicolumn{3}{c}{\textbf{Highest Uncertainty}}\\
Word & Entropy & Count \\ 
\midrule
24pm             & -16.0789 & 336 \\
labwork          & -16.0749 & 272 \\
colonial         & -16.0689 & 198 \\
zoysn            & -16.0601 & 269 \\
ht               & -16.0522 & 515 \\
txcf             & -15.9982 & 112 \\
arrangements     & -15.9794 & 407 \\
parvus           & -15.9773 & 132 \\
nas              & -15.9163 & 251 \\
anesthesiologist & -15.8796 & 220 \\

\bottomrule
\end{tabular}
}
\end{sc}
\end{scriptsize}
\end{center}
\end{table} 
Another motivation for model uncertainty lies in understanding which feature values are most responsible for the variance of the predictive uncertainty distribution.  Our \gls*{rnn} with Bayesian embeddings model is particularly well suited for this task in that the uncertainty in embedding space directly corresponds to the predictive uncertainty distribution and represents uncertainty associated with the discrete feature values.  Understanding model uncertainty associated with features can provide some level of interpretability by allowing us to recognize particularly difficult examples and understand which feature values are leading to the disagreement amongst models.  Additionally, it provides a means of determining the types of patient examples that could be beneficial to add to the training dataset for future updates to the model.

For this analysis, we focus on the free-text clinical notes found in the  \gls*{ehr}. For each word in the notes vocabulary, we have an associated embeddings distribution formulated as a multivariate normal distribution. We rank each word by its level of model uncertainty, which we measure in this case by the entropy of its embedding distribution. \autoref{tab:notes-entropy-freq} lists the top and bottom ten words, along with each word's count in the training dataset.  We find, in general, that common words, both subjectively and based on prevalence counts, have lower entropy and thus limited model uncertainty, while rarer words have higher entropy levels, which corresponds to higher model uncertainty.
However, there is a nonlinear relationship between prevalence and entropy, which can be seen, for example, with the word "tracing", which has approximately a 25\% lower count than the other nine words in the bottom ten words, yet has the fifth lowest entropy. This provides some evidence that the model uncertainty is context-specific.

\begin{figure}[ht]
\begin{center}
\begin{center}
\includegraphics[width=\columnwidth]{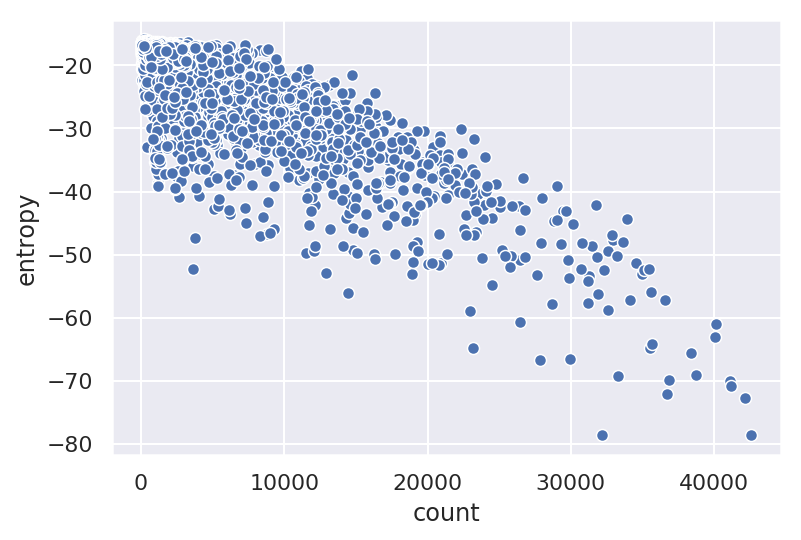}
\end{center}
\caption{Correlation between the entropy of the Bayesian embedding distributions for free-text clinical notes and the associated word frequency.  We find that rarer words are associated with higher model uncertainty, with a non-trivial level of variance at a given frequency.}
\label{fig:notes-entropy-freq}
\end{center}
\end{figure}

We additionally measure the correlation between entropy and word frequency as visualized in \autoref{fig:notes-entropy-freq}.  We find further confirmation that rarer words are generally associated with higher model uncertainty, but that there is a nonlinear relationship between the two entities.

 \section{Conclusion}

In this work, we demonstrated the need for capturing model uncertainty in medicine and examined methods to do so. Our experiments showed multiple findings.
For example, an ensemble of deterministic \glspl*{rnn} captured individualized uncertainty
that led to high predictive disagreement for some patients, all while the models each maintained nearly equivalent clinically-relevant dataset-level metrics.
Furthermore, this disagreement propagated forward as disagreement over the optimal decision for a given patient.
Significant variability in patient-specific predictions and decisions can be an indicator of when a model decision is likely to be brittle, and it provides an opportunity to identify and collect additional data that could reduce the level of model uncertainty.
As another example, we found that models need only be uncertain around the embeddings for competitive performance, as seen by the \gls*{rnn} with Bayesian embeddings. This provided an additional benefit of enabling the ability to determine the level of model uncertainty associated with individual feature values, allowing for some level of interpretability. Furthermore, using model uncertainty methods, we examined patterns in uncertainty across patient subgroups, showing that models can exhibit higher levels of uncertainty for certain groups.

Future work includes designing more specific and clinically-relevant decision cost functions based on both quantified medical ethics \citep{gillon1994medical} and monetary axes; making optimal decisions in light of both data and model uncertainty; and exploring methods to reduce model uncertainty at both training and prediction time. 
\clearpage
\bibliographystyle{ACM-Reference-Format}
\bibliography{refs} 
\nocite{abadi2016tensorflow, kingma2014adam}
\clearpage
\appendix
\section{Appendix}

\subsection{Additional Metrics and Statistics}
\begin{figure}[hb!]
\begin{center}
\includegraphics[width=\columnwidth]{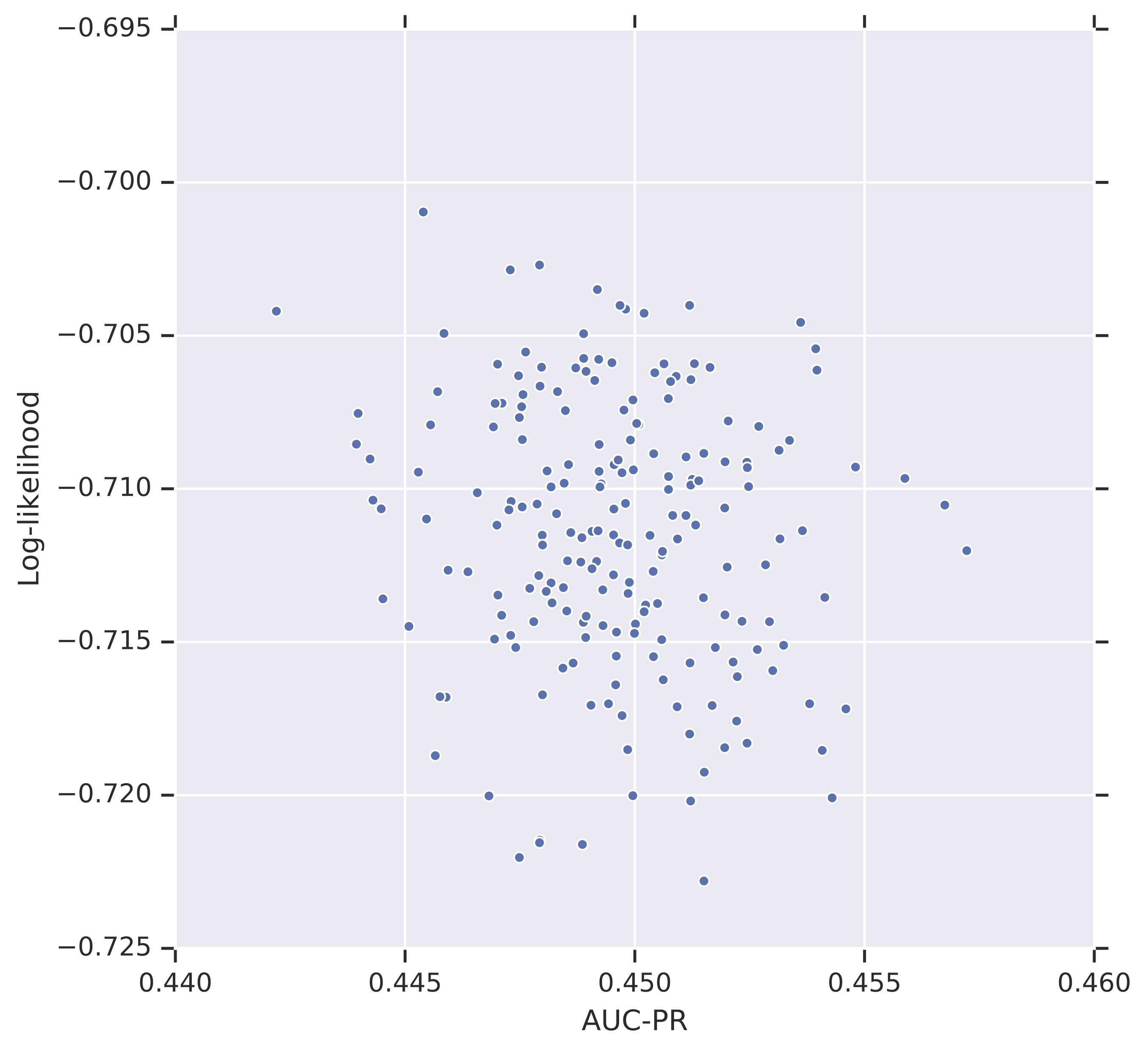}
\end{center}
\caption{Validation \gls*{aucpr} versus held-out log-likelihood values for the deterministic \gls*{rnn} ensemble on the mortality task.  We find that there is no apparent correlation between the two metrics, likely due to the limited differences between the models.}
\label{fig:log-prob-aucpr}
\end{figure}

In \autoref{fig:log-prob-aucpr}, we examine the correlation between held-out log-likelihood and \gls*{aucpr} values for models in the deterministic \gls*{rnn} ensemble on the mortality task.

In \autoref{tab:marginalized-calibration}, we measure the calibration of marginalized predictions of our deterministic \gls*{rnn} ensemble and the Bayesian \glspl{rnn} on the \gls*{mimic} mortality task. We find that the models are all well-calibrated, and that marginalization slightly decreases the calibration error.

\subsection{Additional Training Details}

In terms of hyperparameter optimization, we searched over the hyperparameters listed in \autoref{tab:hyp-search} for the original deterministic \gls*{rnn} (all others in the ensemble differ only by the random seed) and each of the Bayesian models.  \autoref{tab:hyp-values} lists the final hyperparameters associated with each of the models presented in the paper.

Models were implemented using TensorFlow 2.0 \cite{abadi2016tensorflow}, and trained on machines equipped with Nvidia's V100 using the Adam optimizer \cite{kingma2014adam}. MIMIC-III and eICU datasets were each split into train, validation, and test sets in 8:1:1 ratios.

\begin{table*}
\caption{Calibration error for marginalized predictions on the mortality task for an average over $M = 200$ models in the deterministic \gls*{rnn} ensemble, and $M = 200$ samples from each of the Bayesian \gls*{rnn} models.  We find that marginalization slightly increases the calibration of the deterministic ensemble, and that the Bayesian models are comparably well-calibrated.}
\label{tab:marginalized-calibration}
\begin{center}
\begin{sc}
\begin{tabular}{l | c c c c}
\toprule
Model
& \begin{tabular}{@{}c@{}} Val.\\\,\,\acrshort*{ece} $\downarrow$\end{tabular}
& \begin{tabular}{@{}c@{}} Val.\\\,\,\acrshort*{ace} $\downarrow$\end{tabular}
& \begin{tabular}{@{}c@{}} Test\\\,\,\,\acrshort*{ece} $\downarrow$\end{tabular}
& \begin{tabular}{@{}c@{}} Test\\\,\,\,\acrshort*{ace} $\downarrow$\end{tabular} \\
\midrule
\begin{tabular}{@{}l@{}}Deterministic Ensemble\end{tabular}
& 0.0157
& 0.0191
& 0.0157
& 0.0191 \\

\begin{tabular}{@{}l@{}}Bayesian Embeddings\end{tabular}
& 0.0167
& 0.0194
& 0.0163
& 0.0221 \\

\begin{tabular}{@{}l@{}}Bayesian Output\end{tabular}
& 0.0263
& 0.0217
& 0.0241
& 0.0279 \\

\begin{tabular}{@{}l@{}}Bayesian Hidden+Output\end{tabular}
& 0.0194
& 0.0212
& 0.0173
& 0.0240 \\

\begin{tabular}{@{}l@{}}Bayesian RNN+Hidden+Output\end{tabular}
& 0.0240
& 0.0228
& 0.0182
& 0.0247 \\

\begin{tabular}{@{}l@{}}Fully Bayesian\end{tabular}
& 0.0226
& 0.0192
& 0.0178
& 0.0197 \\
\bottomrule
\end{tabular}
\end{sc}
\end{center}
\end{table*}

\begin{table*}[hb]
\caption{Hyperparameters and their associated search sets or ranges.}
\label{tab:hyp-search}
\begin{center}
\begin{sc}
\begin{tabular}{l l}
\toprule
Hyperparameter & Range/Set  \\
\midrule
Batch size & \{32, 64, 128, 256, 512\} \\
Learning rate & [0.00001, 0.1] \\
KL or regularization annealing steps & [1, 1e6] \\
Prior standard deviation (Bayesian only) & [0.135, 1.0] \\
Dense embedding dimension & \{16, 32, 64, 100, 128, 256, 512\} \\
Embedding dimension multiplier & [0.5, 1.5] \\
\gls*{rnn} dimension & \{16, 32, 64, 128, 256, 512, 1024\} \\
Number of \gls*{rnn} layers & \{1, 2, 3\} \\
Hidden affine layer dimension & \{0, 16, 32, 64, 128, 256, 512\} \\
Bias uncertainty (Bayesian only) & \{True, False\} \\
\bottomrule
\end{tabular}
\end{sc}
\end{center}
\end{table*}

\begin{table*}[hb]
\caption{Model-specific hyperparameter values.}
\label{tab:hyp-values}
\begin{center}
\begin{sc}
\resizebox{\textwidth}{!}{
\begin{tabular}{l | c c c c c c c c c c}
\toprule
Model
& \begin{tabular}{@{}c@{}} Batch\\size\end{tabular}
& \begin{tabular}{@{}c@{}} Learning\\rate\end{tabular}
& \begin{tabular}{@{}c@{}} Annealing\\steps\end{tabular}
& \begin{tabular}{@{}c@{}c@{}} Prior\\std.\\dev.\end{tabular}
& \begin{tabular}{@{}c@{}c@{}} Dense\\embedding\\dim.\end{tabular}
& \begin{tabular}{@{}c@{}c@{}} Embeddimg\\dim.\\multiplier\end{tabular}
& \begin{tabular}{@{}c@{}} \Gls*{rnn}\\dim.\end{tabular}
& \begin{tabular}{@{}c@{}c@{}} Num.\\\gls*{rnn}\\layers\end{tabular}
& \begin{tabular}{@{}c@{}c@{}} Hidden\\layer\\dim.\end{tabular}
& \begin{tabular}{@{}c@{}} Bias\\uncertainty\end{tabular} 
\\

\midrule
Deterministic Ensemble
& 256
& 3.035e-4
& 1
& --
& 32
& 0.858
& 1024
& 1
& 512
& -- 
\\

Bayesian Embeddings
& 256
& 1.238e-3
& 9.722e+5
& 0.292
& 32
& 0.858
& 1024
& 1
& 512
& False 
\\

Bayesian Output
& 256
& 1.647e-4
& 8.782e+5
& 0.149
& 32
& 0.858
& 1024
& 1
& 512
& False
\\

Bayesian Hidden+Output
& 256
& 2.710e-4
& 9.912e+5
& 0.149
& 32
& 0.858
& 1024
& 1
& 512
& False
\\

Bayesian RNN+Hidden+Output
& 512
& 1.488e-3
& 6.342e+5
& 0.252
& 32
& 1.291
& 16
& 1
& 0
& True
\\

Fully Bayesian
& 128
& 1.265e-3
& 9.983e+5
& 0.162
& 256
& 1.061
& 16
& 1
& 0
& True
\\
\bottomrule
\end{tabular}
}
\end{sc}
\end{center}
\end{table*} 
\end{document}